\documentclass{article}

\usepackage[main, final]{neurips_2025}
\usepackage[utf8]{inputenc} % allow utf-8 input
\usepackage[T1]{fontenc}    % use 8-bit T1 fonts
\usepackage{hyperref}       % hyperlinks
\usepackage{url}            % simple URL typesetting
\usepackage{booktabs}       % professional-quality tables
\usepackage{amsfonts}       % blackboard math symbols
\usepackage{nicefrac}       % compact symbols for 1/2, etc.
\usepackage{microtype}      % microtypography
\usepackage{xcolor}         % colors
\usepackage{amsmath}
\usepackage[capitalize]{cleveref}
\usepackage{bm}
\usepackage{wrapfig}

\crefname{section}{\S}{\S\S}
\crefname{subsection}{\S}{\S\S}
\crefname{conj}{Conj.}{Conj.}

\Crefname{assumption}{\textbf{H}\hspace{-3pt}}{\textbf{H}\hspace{-3pt}}
\crefname{assumption}{\textbf{H}}{\textbf{H}}

\crefname{algorithm}{\text{Alg.}}{\text{Alg.}}
\crefname{assumption}{\textbf{H}}{\textbf{H}}
\crefname{equation}{\text{Eq}}{\text{Eq}}
\crefname{definition}{\text{Dfn.}}{\text{Dfn.}}
\crefname{lemma}{\text{Lemma}}{\text{Lemma}}
\crefname{dfn}{\text{Dfn.}}{\text{Dfn.}}
\crefname{thm}{\text{Thm.}}{\text{Thm.}}
\crefname{tab}{\text{Tab.}}{\text{Tab.}}
\crefname{fig}{\text{Fig.}}{\text{Fig.}}
\crefname{table}{\text{Tab.}}{\text{Tab.}}
\crefname{figure}{\text{Fig.}}{\text{Fig.}}

\usepackage{arydshln}
\usepackage{multirow}
\usepackage{graphicx}
\usepackage{subcaption}
\usepackage{xspace}

\newcommand{\eg}{\textit{e.g.}\xspace}
\newcommand{\parahead}[1]{\vspace{1mm}\noindent\textbf{{#1}.}\ }

\newcommand{\methodname}{{BTimer}}
\newcommand{\enhancer}{{NTE}}

\usepackage{ifthen}
\newboolean{showcomments}
\setboolean{showcomments}{false} % Set to true to show comments, false to hide

\newcommand{\jhc}[1]{\ifthenelse{\boolean{showcomments}}{{\color{red}Jiahui: {#1}}}{}}
\newcommand{\ZG}[1]{\ifthenelse{\boolean{showcomments}}{{\color{blue}Zan: {#1}}}{}}
\newcommand{\HX}[1]{\ifthenelse{\boolean{showcomments}}{{\olive{blue}HX: {#1}}}{}}
\newcommand{\JR}[1]{\ifthenelse{\boolean{showcomments}}{{\cyan{blue}JR: {#1}}}{}}
\newcommand{\AM}[1]{\ifthenelse{\boolean{showcomments}}{{\cyan{orange}AM: {#1}}}{}}

\usepackage{colortbl}
\usepackage{xcolor}
\usepackage{siunitx}
\definecolor{first}{rgb}{1, 0.7, 0.7}
\definecolor{second}{rgb}{1,0.85, 0.7}
\definecolor{third}{rgb}{1,1, 0.8}

\newcommand{\Real}{{\mathbb{R}}}
\newcommand{\SE}{{\mathbb{SE}(3)}}

\newcommand{\Video}{{\mathcal{I}}}
\newcommand{\PoseSet}{{\mathcal{P}}}
\newcommand{\TimeSet}{{\mathcal{T}}}
\newcommand{\ImageSet}{{\mathcal{I}}}
\newcommand{\Gaussian}{{\mathbf{G}}}
\newcommand{\GaussianSet}{{\mathcal{G}}}
\newcommand{\Image}{{\mathbf{I}}}
\newcommand{\Pose}{{\mathbf{P}}}
\newcommand{\Loss}{{\mathcal{L}}}

\newcommand{\Embedding}{{\mathbf{e}}}
\newcommand{\Feature}{{\mathbf{f}}}

\makeatletter
\DeclareRobustCommand\onedot{\futurelet\@let@token\@onedot}
\def\@onedot{\ifx\@let@token.\else.\null\fi\xspace}

\def\eg{\emph{e.g}\onedot} 
\def\ie{\emph{i.e}\onedot} 
\def\cf{\emph{cf}\onedot}

\makeatother

\title{Feed-Forward Bullet-Time Reconstruction of Dynamic Scenes from Monocular Videos}

\author{%
  Hanxue Liang$^{1,2}$\footnotemark[1], Jiawei Ren$^{1,3}$\footnotemark[1], Ashkan Mirzaei$^{1,4}$\footnotemark[1], Antonio Torralba$^{1,5}$,  \\ \textbf{Ziwei Liu$^{3}$}, 
  \textbf{Igor Gilitschenski$^{4}$}, \textbf{Sanja Fidler$^{1,4,6}$},  \\\textbf{Cengiz Oztireli$^{2}$}, \textbf{Huan Ling$^{1,4,6}$}, \textbf{Zan Gojcic$^{1}$}\footnotemark[2], \textbf{Jiahui Huang$^{1}$}\footnotemark[2] \\ \\
  $^{1}$NVIDIA, $^{2}$University of Cambridge, $^{3}$Nanyang Technological University, \\ $^{4}$University of Toronto, $^{5}$MIT, $^{6}$Vector Institute \\
  \url{https://research.nvidia.com/labs/toronto-ai/bullet-timer/}\\
}

\begin{document}

\maketitle

\begingroup
\renewcommand\thefootnote{}\footnotetext{%
  \hspace{-0.5em}$^*$/$^\dagger$: Equal contribution/advising.%
}%
\endgroup

%\newcommand\blfootnote[1]{%
%  \begingroup
%  \renewcommand\thefootnote{}%
%  \footnotetext{#1}%
%  \addtocounter{footnote}{-1}%
%  \endgroup
%}
%\blfootnote{$^*$/$^\dagger$: Equal contribution/advising.}

%\begin{abstract}
  %The abstract paragraph should be indented \nicefrac{1}{2}~inch (3~picas) on both the left- and right-hand margins. Use 10~point type, with a vertical spacing (leading) of 11~points.  The word \textbf{Abstract} must be centered, bold, and in point size 12. Two line spaces precede the abstract. The abstract must be limited to one paragraph.
%\end{abstract}

%\section{Submission of papers to NeurIPS 2025}
\begin{abstract}
\vspace{-0.5em}
Recent advancements in static feed-forward scene reconstruction have demonstrated significant progress in high-quality novel view synthesis. However, these models often struggle with generalizability across diverse environments and fail to effectively handle dynamic content.
We present \methodname{} (short for \underline{B}ullet\underline{Timer}), \textbf{the first motion-aware feed-forward model for real-time reconstruction and novel view synthesis of dynamic scenes}. 
Our approach reconstructs the full scene in a 3D Gaussian Splatting representation at a given target (`bullet') timestamp by \emph{aggregating} information from all the context frames. 
% This allows training in a fully self-supervised manner without explicit motion annotations.
Such a formulation allows \methodname{} to gain scalability and generalization by leveraging both static and dynamic scene datasets.
% while supporting high-resolution images with a large number of views. 
% We curate a small but high-quality multi-view dataset of dynamic scene datasets, showing that fine-tuning on this subset significantly improves 4D reconstruction performance.
% In the meanwhile, we naturally enable scaling up and hence generalizability by enabling the use of both static and dynamic scene datasets, as well as training and testing on high-resolution images and a large number of views.
% We curate a small subset of high-quality dynamic scene datasets from various sources and demonstrate that finetuning from this subset significantly improves the results.
Given a casual monocular dynamic video, \methodname{} reconstructs a \emph{bullet-time}\footnote{In this paper, we define \emph{bullet-time} as the instantiation of a 3D scene \emph{frozen} at a given/fixed timestamp $t$.} scene within 150ms %\textcolor{red}{on 256 × 256} 
resolution while reaching state-of-the-art performance on both static and dynamic scene datasets, even compared with optimization-based approaches.
% \textbf{Code and pretrained models will be released.}
\end{abstract}    
% \begin{figure}[ht!]
%     \centering
%     \includegraphics[width=0.6\linewidth]{figs/lpips_blobs_v6.pdf}
%     \caption{\textbf{Rendering quality vs. speed.} Our model can reconstruct and render dynamic scenes at a much faster speed than existing approaches with a competitive quality. Numbers are reported on NVIDIA Dynamic Scene Dataset~\cite{yoon2020novel}}
%     \label{fig:speed-quality-chart}
%     \vspace{-10pt}
% \end{figure}

\begin{wrapfigure}{r}{0.5\textwidth}
  \begin{center}
    \vspace{-1.0em}
    \includegraphics[width=\linewidth]{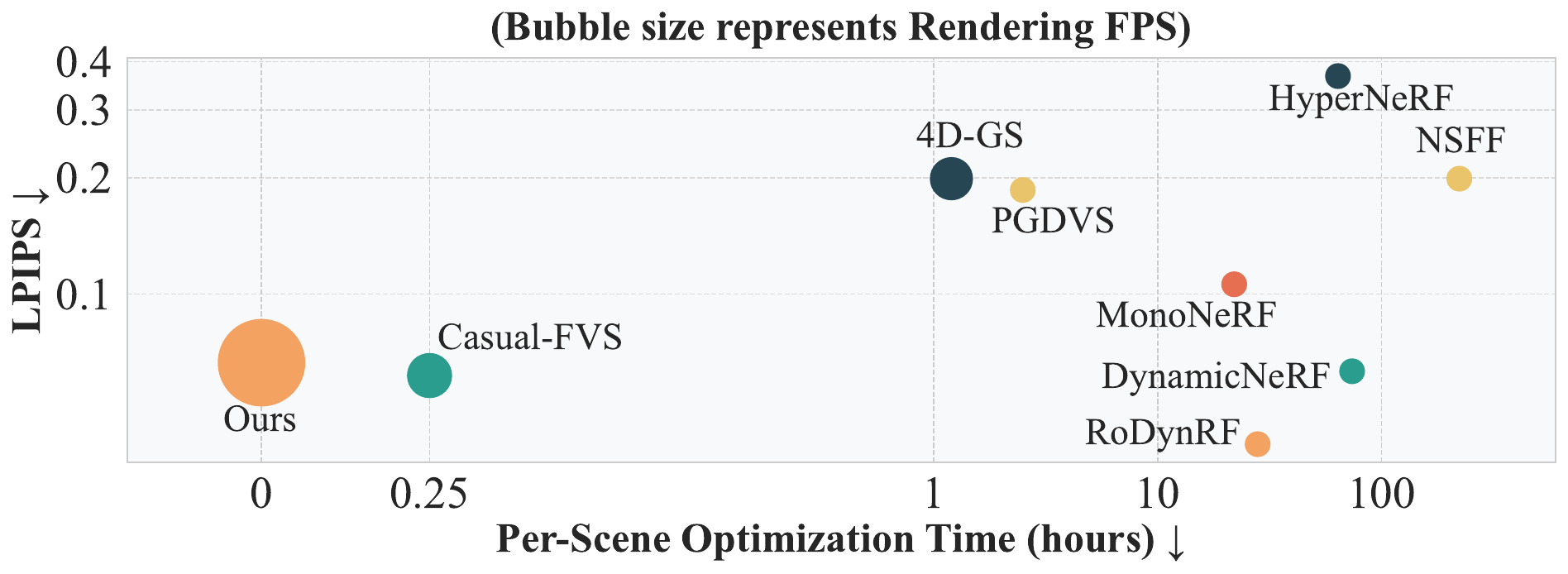}
  \end{center}
    \caption{\textbf{Rendering quality vs. speed.} Our model can reconstruct and render dynamic scenes at a much faster speed than existing approaches with a competitive quality. Numbers are reported on NVIDIA Dynamic Scene Dataset~\cite{yoon2020novel}}
        \label{fig:speed-quality-chart}
        \vspace{-1.0em}
\end{wrapfigure}

\section{Introduction}
\label{sec:intro}

% [Motivation and Challenges]
Multi-view reconstruction and novel-view synthesis are long-standing challenges in computer vision, with numerous applications ranging from AR/VR to simulation and content creation. 
While significant progress has been made in reconstructing static scenes, dynamic scene reconstruction from monocular videos remains challenging due to the inherently ill-posed nature of reasoning about dynamics from limited observations \cite{li2021neural}.

% [Overview of 3D/4D reconstruction models]
Current methods for static scene reconstruction can be broadly divided into two categories: optimization-based~\cite{mildenhall2021nerf,park2021hypernerf} and feed-forward~\cite{zhang2025gs,tian2023mononerf} approaches. However, extending both of these to \emph{dynamic scenes} is not straightforward. To reduce the ambiguities of scene dynamics, many optimization-based methods aim to constraint the problem with data priors such as depth and optical flow \cite{li2021neural,xian2021space,li2023dynibar,lee2025fast}. However, balancing these priors with the data remains challenging~\cite{wang2024shape,lei2024mosca}. 
Moreover, per-scene optimization is time-consuming and thus difficult to scale.

% [Feed-forward models and problems with dynamic models]
On the other hand, to avoid the lengthy per-scene-optimization, recent feed-forward approaches~\cite{hong2023lrm,cong2023enhancing,chen2025mvsplat,charatan2024pixelsplat, jin2024lvsm, zhang2025gs} explored learning generalizable models on large-scale datasets to directly perform static scene reconstructions, thereby learning strong priors from data. %On the other hand, learning-based approaches~\cite{hong2023lrm,cong2023enhancing,chen2025mvsplat,charatan2024pixelsplat,ren2024l4gm} are trained on large-scale datasets to directly predict reconstructions in a feed-forward manner, thereby learning strong priors from data. 
These inherent priors could help resolve ambiguities due to complex motion, but none of previous approaches have yet been extended to dynamic scenes. This limitation stems from both the complexity of modeling dynamic scenes and the lack of 4D supervision data. 
% For effective dynamic reconstruction, a feed-forward model must learn to disentangle scene dynamics from camera motion and reason about complex motions to render the scene at any desired timestamp and viewpoint.
The only feed-forward dynamic reconstruction model~\cite{ren2024l4gm} is thus trained on synthetic object-centric datasets, requires fixed camera viewpoints and multiview supervision, and cannot generalize to real-world scene scenarios.

In this work, we aim to answer the question: \emph{How can one build a feed-forward reconstruction model that can handle dynamic scenes effectively?}
We build upon the recent success of the pixel-aligned 3D Gaussian Splatting (3DGS \cite{kerbl20233d}) prediction models~\cite{zhang2025gs} and propose a novel \emph{bullet-time} formulation for feed-forward dynamic reconstruction. 
The core idea is simple yet effective: we add a \emph{bullet-time} embedding to the context (input) frames, indicating the desired timestamp for the output 3DGS representation. 
Our model is trained to aggregate the predictions of context frames to reflect the scenes at the \emph{bullet} timestamp, yielding a spatially complete 3DGS scene. 
This design not only naturally unifies the static and dynamic reconstruction scenarios, but also enables our model to become implicitly motion-aware while learning to capture scene dynamics. In particular, the proposed formulation \textbf{(i)} allows us to pre-train our model on large amounts of \emph{static} scene data, \textbf{(ii)} scales effectively across datasets, without being constrained by duration and frame rates of input videos, and \textbf{(iii)} outputs volumetric video representations that inherently support multiple viewpoints. Meanwhile, in the presence of fast motions, we additionally introduce a Novel Time Enhancer (\enhancer) module to predict the intermediate frames before feeding them to the main model.
%\HX{do we need to mention that bullet-time recon allows us to freely select distant frames and incorporate multi-view information, which I think is one big advantage}\ZG{I am not fully sure what you mean with this, could you elaborate?}

% [Contributions]
In summary, we present \methodname{}, \emph{the first feed-forward model for real-time reconstruction and novel view synthesis of dynamic scenes.} To achieve this goal, we introduce the core bullet-time formulation and develop a curriculum training strategy that enables the learning of a highly generalizable model on a large, carefully curated dataset comprising both static and dynamic scenes. Furthermore, we present an additional NTE module to effectively handle fast motions, enhancing the model's robustness in challenging scenarios.
% We observed that a model pretrained on large amounts of static scenes could learn to handle dynamics effectively from a smaller set of high-quality annotated dynamic scenes, and the resulting model generalizes well to unseen datasets.
Our method is highly efficient: feed-forward inference with 12 context frames of $256\times 256$ resolution only costs 150ms on a single GPU, and the output 3DGS can be rendered in real-time. 
%\methodname{} is trained using a large curated dataset comprising both static and dynamic scenes, 
%\methodname{} can handle both indoor and outdoor scenes, and can handle both static and dynamic scenarios 
\methodname{} is capable of handling both static and dynamic reconstructions. It achieves competitive results on various reconstruction benchmarks, even surpassing many expensive per-scene optimization-based methods, as illustrated in \cref{fig:speed-quality-chart}.
%\vspace{-1.0em}
%(maybe we should also include static performance here but not emphasize)
% \methodname{} outperforms all existing static models \HX{what's the comparison with static model here?} and achieves competitive results on dynamic scenes, even when compared to computationally expensive per-scene optimization methods.
\section{Related work}
\label{sec:rw}
\vspace{-0.5em}
\parahead{Dynamic 3D representations}%
Depending on the tasks at hand, typical choices of 3D representations include voxels~\cite{liu2020neural,williams2024fvdb}, implicit fields/NeRFs~\cite{mescheder2019occupancy,mildenhall2021nerf,muller2022instant}, and point clouds/3D Gaussians~\cite{aliev2020neural,kerbl20233d}.
Representing dynamics on top has an even larger design space:
One existing line of works directly builds a `4D' representation to enable feature queries at arbitrary positions and timestamps from an implicit field~\cite{fridovich2023k,cao2023hexplane} or via marginalization at a given step~\cite{yang2023real,duan20244d}, with the extensibility to higher dimensions such as material~\cite{bemana2020x}. Another line of work first defines a canonical 3D space, and learns a deformation field to warp the canonical space to the target frame. %Such a deformation field can be parametrized via embedding graphs~\cite{luiten2023dynamic}, control points~\cite{huang2024sc}, motion scaffolds~\cite{lei2024mosca}, rigid transformation modes~\cite{wang2024shape} or bounding boxes~\cite{zhou2024drivinggaussian,chen2024omnire}, explicit scene flows~\cite{lin2024gaussian}, or implicit flow fields~\cite{wu20244d,yang2024deformable,fang2022fast,ren2023dreamgaussian4d}. 
While these methods learn additional information about shape correspondences, their performance heavily relies on the quality and topology of the canonical space.

\parahead{Dynamic novel view synthesis}
For tasks that require a relatively smaller view extrapolation, the problem of novel view synthesis can be tackled without explicit 3D geometry in the loop, using depth warping~\cite{yoon2020novel} or multi-plane images~\cite{tucker2020single}.
Otherwise, the study of novel view synthesis of dynamic scenes~\cite{pumarola2021d,park2021hypernerf} is mainly on (1) effectively optimizing the 3D representation through input images through monocular cues~\cite{li2022neural,li2023dynibar,lei2024mosca,wang2024shape} or geometry regularizations~\cite{park2021nerfies,liu2023robust}, and (2) being able to render fast with grids~\cite{attal2023hyperreel}, local-planes~\cite{lee2023fast}, or dynamic 3D Gaussians~\cite{wang2024gflow} formulation.
Our method aims to provide a dynamic representation that is fast to build within hundreds of milliseconds while reaching competitive rendering quality as the above optimization-based methods.

\parahead{Feed-forward reconstruction models}
In many applications where the reconstruction speed is crucial, most optimization-based reconstruction methods become less preferable.
To this end,% one line of methods uses data priors to serve as an initialization~\cite{fan2024instantsplat} or guidance~\cite{chen2025g3r,tian2023mononerf}, but they still need few-shot optimization to refine the results~\cite{cong2023enhancing,wang2022attention}.
one line of work that starts to emerge is fully feed-forward models that directly regress from 2D images to 3D, represented as either neural field \cite{cong2023enhancing,wang2022attention}, triplanes~\cite{hong2023lrm}, 3D Gaussians~\cite{chen2025mvsplat,zhang2025gs,charatan2024pixelsplat}, sparse voxels~\cite{ren2024scube}, or latent tokens~\cite{jin2024lvsm}.
Crucially, while feed-forward reconstruction models for static scenes have seen development, the extension to dynamic scenes is still challenging.
Existing methods either require hard-to-acquire consistent video depth as input~\cite{zhao2024pseudo}, do not support rendering~\cite{zhang2024monst3r}, or only work on object-scale data~\cite{ren2024l4gm}.
% PGDVS\cite{zhao2024pseudo} makes the first attempt to do 
%  dynamic novel view synthesis by proposing a pseudo-generalized process without scene-specific appearance
% optimization. But their model needs geometrically and temporally consistent depth estimates  which takes approximately three GPU hours for each video.
% L4GM~\cite{ren2024l4gm} only works on objects, and is requires hallucinated multi-view images as input.
% MonST3R~\cite{zhang2024monst3r} builds upon \cite{wang2024dust3r} to regress non-renderable point clouds only and requires additional optimization step for motion decomposition.
In contrast, our method supports reconstructing from a monocular video containing dynamic scenes in a fully feed-forward manner, and is able to render at arbitrary viewpoints and timestamps.
\vspace{-0.5em}

\begin{figure}
    \centering
    \includegraphics[width=0.85\linewidth]{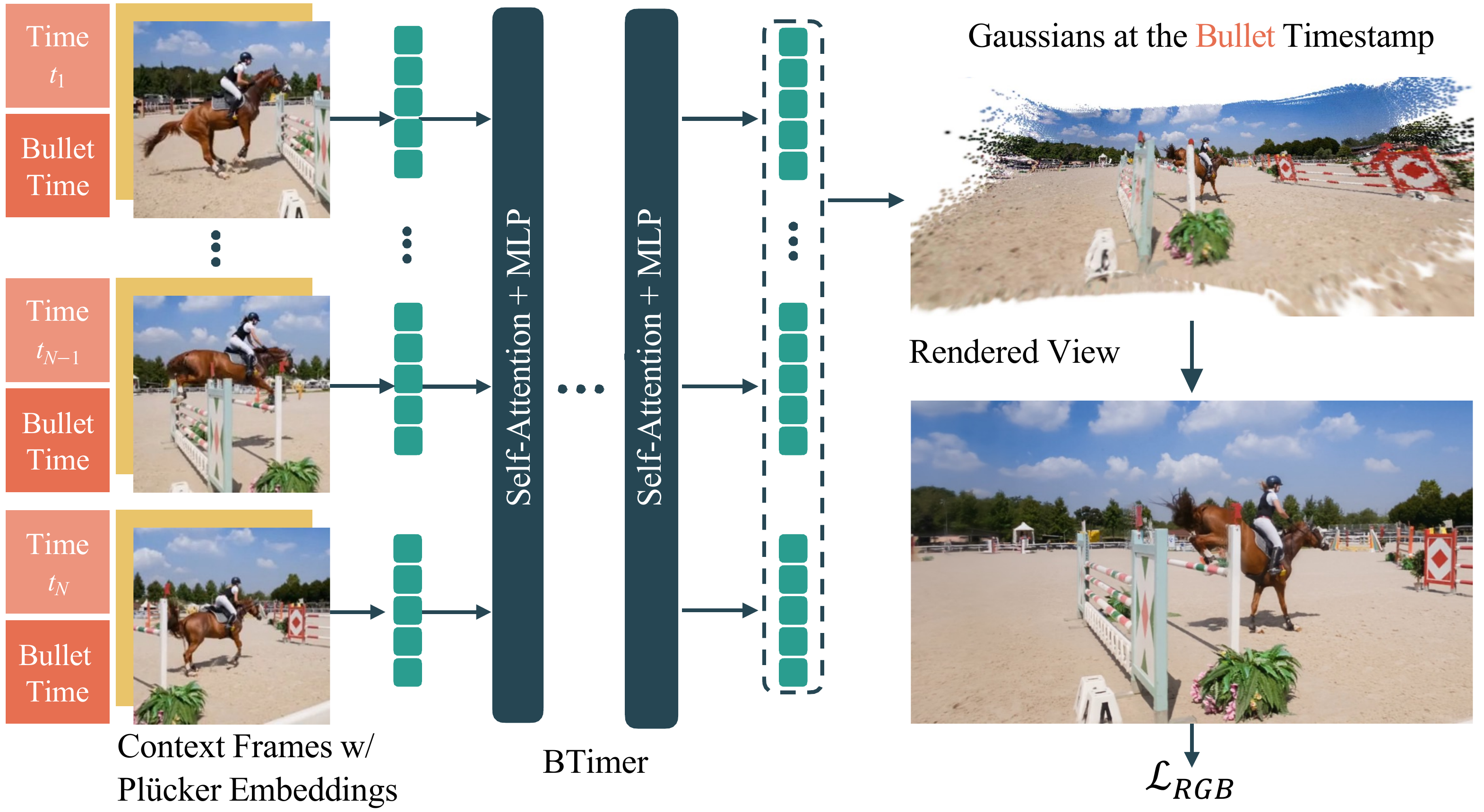}
    \caption{\textbf{\methodname{}.} The model takes as input a sequence of context frames and their Plücker embeddings, along with the context timestamp and target (`bullet') timestamp embeddings. It then directly predicts the 3DGS representation at the bullet timestamp.}
    \vspace{-4mm}
    \label{fig:method}
\end{figure}

\section{Method}
\label{sec:method}

\parahead{Overview}%
Given a monocular video (image sequence) represented by $\Video = \{ \Image_i \in \Real^{H \times W \times 3} \}_{i=1}^N$ with $N$ frames of width $W$ and height $H$, along with known camera poses $\PoseSet = \{ \Pose_i \in \SE \}_{i=1}^N$, intrinsics, and corresponding timestamps $\TimeSet = \{ t_i \in \Real \}_{i=1}^N$, our goal is to build a feed-forward model capable of rendering high-quality novel views at arbitrary timestamps $t \in [t_1, t_N]$. 

The core of our approach is a transformer-based \emph{bullet-time} reconstruction %ViT~\cite{dosovitskiy2020image} 
model, named \textbf{\methodname}, that takes in a subset of frames $\Video_{\text{c}} \subset \Video$ (denoted as \emph{context frames}) along with their corresponding poses $\PoseSet_{\text{c}} \subset \PoseSet$ and timestamps $\TimeSet_{\text{c}} \subset \TimeSet$, and outputs a complete 3DGS~\cite{kerbl20233d} scene frozen at a specified \emph{bullet} timestamp $t_b \in [\min_{\TimeSet_{\text{c}}},\max_{\TimeSet_{\text{c}}}]$ (\cref{sec:bullet-time model}). 
 Iterating over all $t_b \in \TimeSet$ results in a full video reconstruction represented by a sequence of 3DGS.
% reconstructs a static scene at a specific bullet time-step given all video frames and their camera trajectory (\cref{sec:bullet-time model}). 
% Effective time conditioning techniques and training losses are make the model motion-aware.
% We propose time conditioning and surrogate temporal supervision to make the model aware of time between different frames input. 
% 3D Gaussian Splatting is used as a 3D representation for fast rendering.
We further introduce a Novel Time Enhancer (\enhancer) module that synthesizes interpolated frames with timestamps $t \notin \TimeSet$ (\cref{sec:lvsm-4d}). 
The output of the \enhancer{} module is used along with other context views as input to the bullet-time model to enhance reconstruction at arbitrary intermediate timestamps.
% To enhance reconstruction at \emph{arbitrary} intermediate timestamps $t \notin \TimeSet$, \HX{how about: we introduce a novel time interpolation model that synthesizes frames beyond the provided timestamps. (to avoid wrong impression that we use another model to do novel view synthesis.) }we introduce an interpolation model that synthesizes novel views beyond the provided poses and timestamps (\cref{sec:lvsm-4d}). 
% Specifically, the interpolation model generates frames at the desired timestamp $t$, which are then used in conjunction with other context views as input to the bullet-time model. 
% \ZG{Here I think that we need to be more clear what is the difference of the two models, and why don't we then just use the interpolation model for everything} \JR{How about: Note that, unlike BulletTimer, the interpolation model does not output a 3D representation, and we empirically find it less capable of view extrapolation.}
To effectively train our model, we carefully design a learning curriculum (\cref{sec:3d-pretrain}) that incorporates a large mixture of datasets containing both static and dynamic scenes, to enhance motion awareness and temporal consistency of our models.
% Due to the scarcity of 4D data, we pretrain our models on a curated selection of \emph{static} datasets spanning various scales and domains, using a carefully designed curriculum (\cref{sec:3d-pretrain}). This pretraining is followed by fine-tuning on a limited set of high-quality 4D data to enhance motion awareness and temporal consistency of our models.

\subsection{\methodname{} reconstruction model}
\label{sec:bullet-time model}

%Our bullet-time reconstruction model takes a collection of images $\ImageSet$ and their corresponding camera poses $\PoseSet$ and timestamps $\TimeSet$ as input. The output is a 3DGS scene at a the \emph{bullet} time $t_b \in \TimeSet$. With a trained model, we could iterate over all the video timestamps and obtain a full dynamic reconstruction. \ZG{I would completely remove this part as it is repetition of the above.}

% We divide and conquer the dynamic reconstruction problem into static reconstruction at multiple time steps. 
% We propose a feed-forward bullet-time reconstruction that takes a sequence of video frames and a target frame indicator as input and outputs the 3D scene at the targeted time. 
% Once the model is trained, we can parallelly apply the model on multiple time steps to reconstruct a full volumetric video.

\parahead{Model design}% 
%Inspired by \cite{zhang2025gs}, we use a transformer-based model as our backbone. 
Inspired by \cite{zhang2025gs}, our \methodname{} model uses a ViT-based~\cite{dosovitskiy2020image} network as its backbone, consisting of 24 self-attention blocks with LayerNorms~\cite{ba2016layer} applied at both the beginning and the end of the model. 
We divide each input context frame $\Image_i \in \Video_{\text{c}}$ into $8 \times 8$ patches, which are projected into feature space $\{ \Feature_{ij}^{\text{rgb}} \}_{j=1}^{HW/64}$ using a linear embedding layer. The camera Plücker embeddings~\cite{xu2023dmv3d} derived from the camera poses $\Pose_i \in \PoseSet_{\text{c}}$ and the time embeddings (detailed later) are processed similarly to form the camera pose features $\{ \Feature_{ij}^{\text{pose}} \}$ and the time features $\{ \Feature_{i}^{\text{time}} \}$ (shared for all patches $j$).
% (Time feature is the same for all patches that belongs to the same  frame).  %\ZG{I assume that the time feature is the same for all patches that belong to a single image?}. 
These features are added together to form the input tokens for the patches of the context frame $ \{\Feature_{ij}\}_{j=1}^{HW/64}$, where $\Feature_{ij} = \Feature_{ij}^{\text{rgb}}+\Feature_{ij}^{\text{pose}}+\Feature_{i}^{\text{time}} $. 
The input tokens from all context frames are concatenated and fed into the Transformer blocks.

Each corresponding output token $\Feature_{ij}^\text{out}$ is decoded into 3DGS parameters  $\Gaussian_{ij} \in \Real^{8 \times 8 \times 12}$ using a single linear layer. Each 3D Gaussian is paramaterized by its RGB color $\mathbf{c} \in \mathbb{R}^3$, scale $\mathbf{s} \in \mathbb{R}^3$, rotation represented as unit quaternion $\mathbf{q} \in \mathbb{R}^4$, opacity $\sigma \in \mathbb{R}$, and ray distance $\tau \in \mathbb{R}$, resulting in 12 paramaters per Gaussian. The 3D position of each Gaussian $\bm{\mu} \in \mathbb{R}^3$ is obtained through pixel-aligned unprojection as $\bm{\mu}=\mathbf{o}+\tau \mathbf{d}$, where $\mathbf{o} \in \mathbb{R}^3$ and $\mathbf{d} \in \mathbb{R}^3$ are the ray origin and direction obtained from $\Pose_i$.

\parahead{Time embeddings}%
\iffalse
We append two time-related embeddings as input to the network: 
\textbf{(i)} \textbf{Context Time} embedding $\{ \Embedding_{i}^{\text{contexttime}} \}$ that encodes the timestamp $t_i$ for each context frame  $\Image_i$, %corresponding to the input frame, \ie along with $\Image_i$ we use the corresponding $t_i$, 
and 
\textbf{(ii)} \textbf{Bullet Time} embedding $\{ \Embedding_{i}^{\text{bullettime}} \}$ that encodes the bullet timestamp $t_b$ that we want to reconstruct. Note, that all input context frames share the \emph{same} bullet time embedding. Both timestamp scalars $t$ are encoded using standard Positional Encoding (PE)~\cite{vaswani2017attention} with sinusoidal functions and then passed through two linear layers to obtain the final features. These two features are added together to form the final time features  $ \Feature_{i}^{\text{time}} = \Feature_{i}^{\text{contexttime}} + \Feature_{i}^{\text{bullettime}}$, which is then added to input tokens. %\ZG{we are a bit inconsistent with the usage of embedding and feature. Let's check and align tha the embedding is before the projection with the linear layers, while the features is the output of those linear layers} \HX{good catch, fixed}
\fi
The aforementioned input time feature $\Feature_i^\text{time}$ is obtained from:
\textbf{(i)} \textbf{context} timestamp $t_i$ that is separate for each context frame  $\Image_i$, %corresponding to the input frame, \ie along with $\Image_i$ we use the corresponding $t_i$, 
and 
\textbf{(ii)} \textbf{bullet} timestamp $t_b$ that is shared across all context frames $i$. 
% Note, that all input context frames share the \emph{same} bullet time embedding. 
Both timestamp scalars are encoded using standard Positional Encoding (PE)~\cite{vaswani2017attention} with sinusoidal functions, and then passed through two linear layers to obtain the features $\Feature_{i}^{\text{ctx}}$ and $\Feature_{i}^{\text{bullet}}$ respectively. 
Finally, we set $ \Feature_{i}^{\text{time}} = \Feature_{i}^{\text{ctx}} + \Feature_{i}^{\text{bullet}}$.

% \begin{figure}
%     \centering
%     \includegraphics[width=0.6\linewidth]{figs/lvsm_v6_rename.pdf}
%     \caption{\textbf{\enhancer{} Module.} It takes as input the target bullet time embedding, target pose, as well as adjacent frames, and directly predicts corresponding RGB values. The predicted frame is then used in \methodname{} as \emph{bullet} frame for novel time reconstruction.}
%     \vspace{-1em}
%     \label{fig:lvsmillustration}
% \end{figure}

\begin{wrapfigure}{r}{0.5\textwidth}
  \begin{center}
    \vspace{-1em}
    \includegraphics[width=\linewidth]{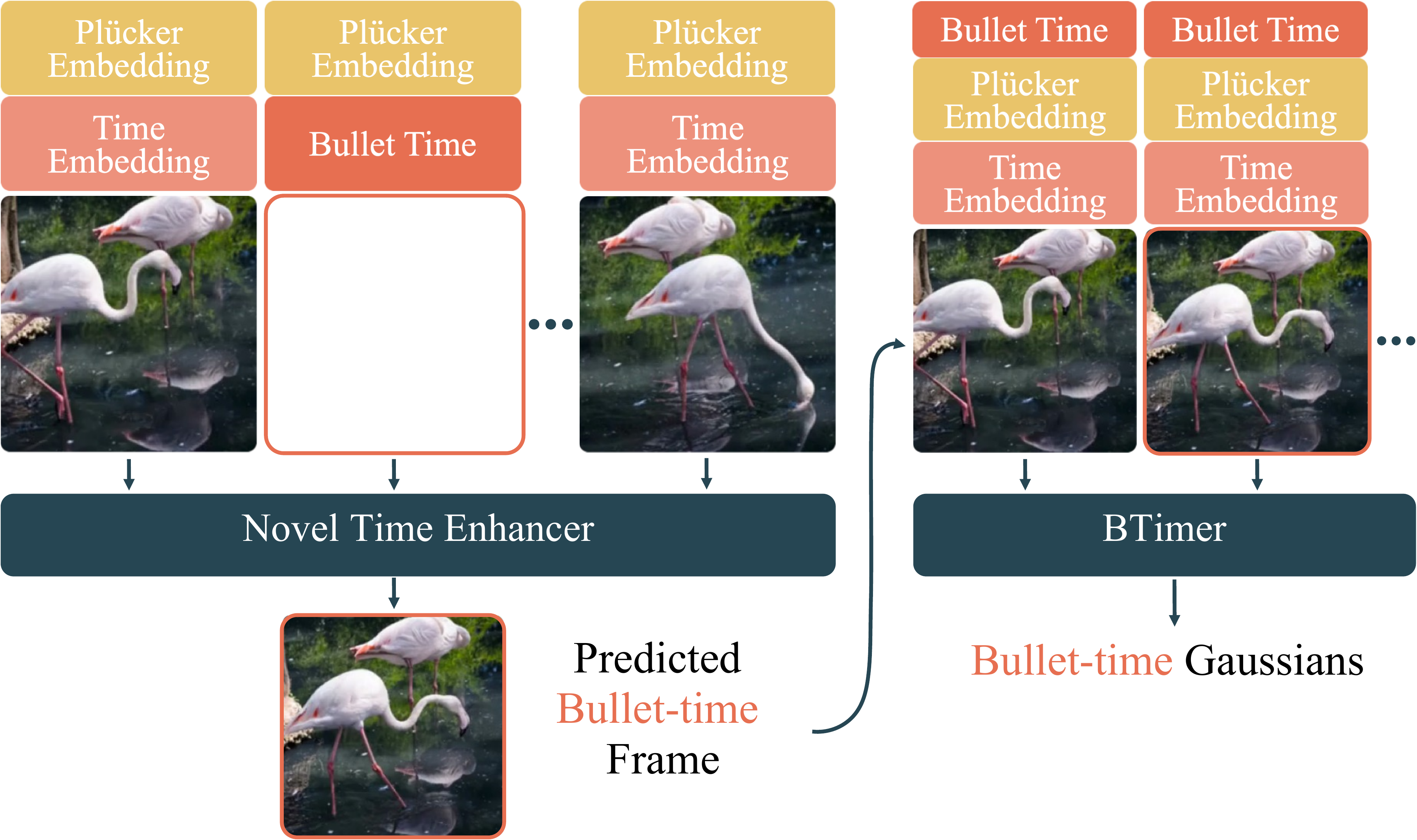}
  \end{center}
\caption{\textbf{\enhancer{} Module.} It takes as input the target bullet time embedding, target pose, as well as adjacent frames to directly predict corresponding RGB values. The predicted frame is then used in \methodname{} as \emph{bullet} frame for novel time reconstruction.}
    \label{fig:lvsmillustration}
\end{wrapfigure}

\parahead{Supervision loss}%
Our model is supervised only by losses defined in the RGB image space, bypassing the need for any source of 3D ground truth that is hard to obtain for real data.
The final loss is a weighted sum of Mean Squared Error (MSE) loss and Learned Perceptual Image Patch Similarity (LPIPS)~\cite{zhang2018perceptual} loss between the images rendered from the 3DGS output and the ground-truth image:
\begin{equation}
    \Loss_\text{RGB} = \Loss_\text{MSE} + \lambda \Loss_\text{LPIPS},
\end{equation}
with $\lambda = 0.5$.

% Following 3D reconstruction model training, we supervise novel-view renderings with ground truth novel-view images. 
% We apply both mean square error on RGB and LPIPS loss as follows:
Careful selection of input context frames and corresponding supervision frames (at the bullet timestamp) during training is essential for stable training and good convergence. In practice, we find the combination of the following two strategies particularly effective:
\textbf{(i)} \textbf{In-context Supervision} where the supervision timestamp is randomly selected from the context frames, encouraging the model to accurately localize and reconstruct the context timestamps. 
For multi-view video datasets, images from additional viewpoints can also contribute to the loss.
\textbf{(ii)} \textbf{Interpolation Supervision} where the supervision timestamp lies between two adjacent context frames. %, but at a timestamp for which we have supervision. 
This forces the model to interpolate the dynamic parts while maintaining consistency for the static regions. 
The interpolation supervision significantly impacts our final performance (\cf \cref{sec:ablation} for details); without it, the model falls into a local minima by positioning the 3D Gaussians close to the context views but hidden from other views. %, to overfit the in-context supervision. 
% This workaround leads to white-edge artifacts in target views, an issue that becomes particularly pronounced when multi-view supervision is unavailable during 4D training (see \cref{sec:ablation} for details).
% Further information on how we leverage existing static-scene datasets for pretraining since existing 4D data are limited in size, is provided in \cref{sec:3d-pretrain}. %\ZG{we are inconsistent with cref and section ref}

% Besides, we propose to leverage single-view videos with annotated camera poses. We design the following surrogate training objectives:
% 1) reference-view supervision, which sets the target time to one of the input frame, and computes the photometric loss with the target frame. This encourages the model to focus only on the specified time step.
% 2) Interpolation supervision, which sets the target timesteps somewhere between input timesteps. A photometric loss is computed between the rendered novel view and the ground truth frame. This challenges the model to reconstruct the static parts while interpolating dynamic parts.
% 3) depth supervision, we compute the normalized depth rendered from the 3D GS with the ground truth depth. 
% The surrogate objective only requires camera annotation, which can potentially scale to internet-scale data.
% However 4D data, \ie, multi-view videos, are limited. One remedy is to co-train with static multiview data, which we elaborate in \cref{sec:3d-pretrain}. 

\parahead{Inference}
%\HX{will reviewer argue about how long the video can be? We can now process 48 frames video. Shall we change to Inference, and just mention our advantage when applying on long video.}
Our \emph{bullet-time} formulation makes it straightforward to reconstruct a full video, which only involves iteratively setting the bullet timestamp $t_b$ to every single timestamp in the video, and can be done efficiently in parallel. 
For a video longer than the number of training context views $|\Video_{\text{c}}|$, at timestamp $t$, apart from including this exact timestamp and setting $t_b = t$, we uniformly distribute the remaining $|\Video_{\text{c}}| - 1$ required context frames across the whole duration of the video to form the input batch with $|\Video_{\text{c}}|$ frames.
%\jhc{I removed the part where we select the nearest context frames to the target timestamp, since this seems to be solved in the next subsection.}

% During inference, reconstructing a 3D scene at different moments only requires changing the time embedding.
% When the video length is larger than the training window, we select T-1 support views that have camera views covering the scene as much as possible. Usually uniformly sampling from the video will be sufficient. We then add the frame closest to the target time into input views, to form a T-frame input batch.

\subsection{Novel time enhancer (\enhancer{}) module} 
\label{sec:lvsm-4d}
While our \methodname{} model can already reconstruct the 3DGS representation for all observed timestamps, we notice that forcing it to reconstruct at a novel intermediate timestamp, \ie performing interpolation at $t_b \notin \TimeSet$, leads to suboptimal results. 
In such cases, the exact bullet-time frame cannot be included in the context frames as it does not exist.
%forcing it to reconstruct at a novel bullet timestamp $t_b$ without video frame, which means $t_b \notin \TimeSet$, will lead to suboptimal results. In this circumstance, we cannot put corresponding frame into context frames as there is no corresponding frame. 
Our model specifically fails to predict a smooth transition between adjacent video frames when the motion is complex and fast. This is mainly caused by the inductive bias of pixel-aligned 3D Gaussian prediction. To mitigate this issue, we propose a \emph{3D-free} Novel Time Enhancer (\enhancer{}) module that directly outputs images at given timestamps, which are then used as input to our \methodname{} model, as illustrated in \cref{fig:lvsmillustration}. %\HX{shall we move this part to combining the models paragraph}While the 3D-free model is good at predicting smooth transitions between adjacent video frames, it struggles to reason about novel views far from the input camera trajectories, as elaborated in ~\cref{sec:ablation}. 

% \ZG{can we add some downside of this model? Like: While the 3D-free model is good at predicting smooth-transitions between two context frames, it struggles to reason about the long term motion ... Is there a result that we can refer to that supports this claim?}

\iffalse
\HX{old version: Empirically, we find that forcing the bullet-time model to reconstruct at a novel bullet timestamp $t_b$ that is not in the context timestamps $t_b \notin \TimeSet$ will lead to suboptimal results.
The model usually fails to predict a smooth transition for the timestamp between adjacent context frames, especially when the motion is complex and fast.
This is mainly restricted by the inductive bias of the image-space representation of the 3DGS scene. We hence supplement a \emph{3D-free} interpolation model that directly outputs images at novel timestamps, which will be used as input to the bullet-time model, illustrated in \cref{fig:lvsmillustration} Figure~\ref{fig:lvsmillustration}.
}
\fi 
% Although the Bullet-time reconstruction model is capable of synthesizing unseen time steps, the interpolation smoothness can be restricted by the pixel-aligned representation. 
% Therefore, we propose a 3D-free view synthesis model that interpolates jointly on camera-trajectory and time. 
% Given a novel time t, we first linearly interpolate the camera trajectory to get the camera pose $c$ at time t. 
% Then we synthesize the novel view at t and c using the model and feed the RGB image and pose as the target view in the Bullet-time model to reconstruct the 3D scene.

\parahead{\enhancer{} module design}%
The design of this module is largely inspired by the very recent decoder-only LVSM~\cite{jin2024lvsm} model. 
Specifically, \enhancer{} copies the same ViT architecture from the \methodname{} model, but the time features of input context tokens only encode their corresponding context timestamps (\ie we set $\Feature_{i}^{\text{time}} = \Feature_{i}^{\text{ctx}}$). 
Additionally, we concatenate extra target tokens to the input tokens, which encode the target timestamp and the target pose for which we want to generate the RGB image. 
Following \cite{jin2024lvsm}, we use QK-norm to stabilize training. 
Implementation-wise we apply an attention mask that masks all the attention to the target tokens, so KV-Cache (\cf \cite{pope2023efficiently}) can be used for faster inference. 
% Input camera poses are normalized so that the distance between the first camera pose and the farthest camera pose is 1, which makes the model invariant to scene scales. 
From the output of the Transformer backbone, we only retain the target tokens, which we then unpatchify and project to RGB values at the original image resolution using a single linear layer.
% \ZG{Jiawei, please check if true}
The interpolation model is trained with the same objective as the main \methodname{} model (see \cref{sec:bullet-time model}), but the output image is directly decoded from the network and not rendered from a 3DGS representation.

\parahead{Integration with \methodname{}}%
While the \enhancer{} module can be used on its own to generate novel views, we empirically find the novel-\emph{view}-synthesis quality to be inferior (\cref{sec:ablation}).
We hence propose to integrate it with our main \methodname{} model.
% While the 3D-free model is good at predicting smooth transitions between adjacent frames, it struggles to reason about novel views far from the input camera trajectories, as elaborated in ~\cref{sec:ablation}. 
% Thus we combine this model with bullet-time reconstruction model to enhance reconstruction performance on videos with fast motion and complex dynamics.
To reconstruct a bullet-time 3DGS at $t_b \notin \TimeSet$, we first use \enhancer{} to synthesize $\Image_b$ at the timestamp $t_b$, where the target pose $\Pose_b$ is linearly interpolated from the nearby context poses in $\PoseSet$, and the context frames are chosen as the nearest frames to $t_b$.
To accelerate the inference of the interpolation model, we use the KV-Cache strategy. In practice we observe that the interpolation model adds negligible overhead to the overall runtime.
\vspace{-1.0em}

\subsection{Curriculum training at scale} 
\label{sec:3d-pretrain}
%\vspace{-2.0em}
One important lesson people have learned from training deep neural networks is to scale up the training~\cite{sutton2019bitter,achiam2023gpt}, and the model's generalizability is largely determined by the data diversity.
Since our bullet-time reconstruction formulation naturally supports both static (by equalizing all elements in $\TimeSet$) and dynamic scenes, and requires only RGB loss for weak supervision, we unlock the potential of leveraging the availability of numerous static datasets to pretrain our model.
% oth \methodname{} and its \enhancer{} module.
% \jhc{NB: We haven't yet specified whether this is for the bullet-time model only or also the interpolation model.}
We hence aim to train a \emph{kitchen-sink} reconstruction model that is \emph{not specific} to any dataset, making it generalizable to both static and dynamic scenes, and capable of handling objects as well as both indoor and outdoor scenes.
This is in contrast to, \eg, GS-LRM~\cite{zhang2025gs} or MVSplat~\cite{chen2025mvsplat} where one needs different models in different domains.

Notably, we apply the following training curriculum to \methodname{} and the \enhancer{} module separately, but during inference they are used jointly as explained in \cref{sec:lvsm-4d}.

%\JR{I merged low-res and high-res static training to stage 1, and add a stage 3 for many-view finetuning.}

\parahead{Stage 1: Low-res to high-res static pretraining}% 
% However, existing 3D reconstruction works often train a model for every single dataset, which often under-performs on out-of-domain data. 
To obtain a more generalizable 3D prior as initialization, we first pretrain the model with a mixture of \emph{static} datasets. 
Time embedding will not be used in this stage. 
The collection of datasets covers object-centric (Objaverse~\cite{deitke2023objaverse}) and indoor/outdoor scenes (RE10K~\cite{zhou2018stereo}, MVImgNet~\cite{yu2023mvimgnet}, DL3DV~\cite{ling2024dl3dv}). 
The datasets cover both the synthetic and real-world domains and consist of ~390K training samples. We normalize the scales of different datasets to be bounded roughly in a $10^3$ cube. 
% Yet, training on datasets that have very different domain styles is still challenging. 
% Therefore, we adopt a curriculum learning strategy. 
Due to the complex data distribution, our training starts from a low-resolution few-view setting that reconstructs on $128 \times 128$ resolution from $|\ImageSet_c| = 4$ context views. 
To further increase the reconstruction details, we fine-tune the model from  $128 \times 128$ by first increasing the image resolution to $256 \times 256$, and then fine-tune to $512 \times 512$. 

\parahead{Stage 2: dynamic scene co-training}% 
After the training on static scenes, we start fine-tuning the model along with time embedding projection layers on dynamic scenes with available 4D data that contains monocular or multi-view synchronized videos.
We leverage Kubric~\cite{greff2022kubric}, PointOdyssey~\cite{zheng2023point}, DynamicReplica~\cite{karaev2023dynamicstereo} and Spring~\cite{mehl2023spring} datasets for  training.
Due to the scarcity of 4D data, during this stage we keep the static datasets for co-training which provides more multi-view supervision and stabilizes the training.
Additionally, we build a customized pipeline to label the camera poses from Internet videos (detailed below), and add them to our training set to further enhance the model's robustness towards real-world data.

\parahead{Stage 3: long-context window fine-tuning}% 
Including more context frames is vital when reconstructing long videos.
Therefore, as a final stage, we increase the number of context views from $|\ImageSet_c| = 4$ to  $|\ImageSet_c| = 12$ to cover more frames. 
Note that this stage does not apply to \enhancer{} as it only takes nearby frames as input.
%\HX{remove the following} When sampling training batches from video dataset, we find it helpful to keep frame gap sufficiently large so that we have enough in-between frames for the interpolation supervision.

\parahead{Annotating internet videos}
We randomly select a subset from the PANDA-70M~\cite{chen2024panda} dataset, and cut the videos into short clips with $\sim$\SI{20}{\second} duration.
We mask out the dynamic objects in the videos with Segment Anything Model~\cite{kirillov2023segment} and then apply DROID-SLAM~\cite{teed2021droid} to estimate the camera poses.
Low-quality videos or annotated poses are filtered out by measuring the reprojection error.
The final dataset contains more than 40K clips with high-quality camera trajectories.
\vspace{-0.8em}
% 4D training data are limited, and we collect them from various sources, with different scene scales and data domains. 
% Fortunately, our models are designed perfectly compatible with static 3D reconstruction. 
% Static reconstruction would be equivalent to setting time steps to the same value across all frames. 
% Therefore, our 4D model can initialize from a 3D pretrained model and co-train with static scene data seamlessly. 

% \parahead{Pre- \& co-training} 
% By training on a collection of 3D datasets at scale, our static model achieves significantly better generalization performance on unseen 3D data.
% We train the static models to initialize the bullet-time model and the 3D-free enhancer separately. 
% During the aforementioned 4D training, we treat static scenes as free-time videos with equal time steps and mix them with dynamic video data for co-training.

% \section{Discussion}
% \subsection{Bullet-time}
% Although the model is also capable of novel view synthesis, it does not produce 3D representation. Moreover, the reconstruction quality is low for far-away views, due to the lack of 3D consistency, and deterrministic.
\section{Experiments}\label{sec:exp}

In this section we first introduce necessary implementation details in \cref{subsec:exp:impl}.
We evaluate the performance of \methodname{} extensively on available dynamic scene benchmarks~\cref{subsec:exp:dynamic}, and demonstrate its \emph{backward} compatibility with static scenes~\cref{subsec:exp:static}.
Ablation studies are found in \cref{sec:ablation}.

\subsection{Implementation details}
\label{subsec:exp:impl}

% \parahead{Implementation Details}
% % Network Configuration and Training / Inference Details
% We use hidden dimension 1024, and time embedding dimension 2. The model has 302M parameters in total. We set the near plane and far plane to 0.1 and 500. We use \textsc{gsplat}~\cite{ye2024gsplatopensourcelibrarygaussian} for efficient GS rasterization, \textsc{FlashAttention-3}~\cite{dao2023flashattention}  and \textsc{FlexAttention}~\cite{he2024flexattention} for attention and masked attention respectively. 

\parahead{Training} 
Our backbone Transformer network is implemented efficiently with FlashAttention-3~\cite{dao2023flashattention} and FlexAttention~\cite{he2024flexattention}.
We use \texttt{gsplat}~\cite{ye2024gsplatopensourcelibrarygaussian} for robust and scalable 3DGS rasterization since the total number of 3D Gaussians generated by our model can be very large.
For \methodname{}, the numbers of training iterations are fixed to 90K, 90K, and 50K for \textbf{Stage 1} training on $128^2$, $256^2$, and $512^2$ resolutions, and are 10K and 5K for \textbf{Stage 2} and \textbf{Stage 3} dynamic scene training respectively. 
We use the initial learning rates of $4 \times 10^{-4}$, $2 \times 10^{-4}$ and $1 \times 10^{-4}$ for the three stages, and apply a cosine annealing schedule to smoothly decay the learning rate to zero.
% The initial learning rate starts from $4e-4$ and will be half for every fine-tuning. 
% The learning rate is cosine annealed to 0 in each round of training. 
% Per-GPU batch size is 8 for $128^2$ and $256^2$, 2 for $512^2$, and 1 for 12-frame $512^2$. 
Training is conducted on 32 NVIDIA A100 GPUs. The learning rate, training GPU numbers and training schedules mainly follow \cite{jin2024lvsm, zhang2025gs}. Training cost analysis and ablation on batch size can be found in the Supplement.
%\textcolor{red}{The full training takes $\sim$4 days on 64 NVIDIA A100 GPUs, which is comparable to existing feed-forward 3D reconstruction methods, such as LVSM~\cite{jin2024lvsm} and LRM~\cite{hong2023lrm} (384 GPU-days) or GS-LRM~\cite{zhang2025gs} (192 GPU-days).} % Note that we train at $480\times270$ and $360\times480$ instead of $512^2$ in dynamic training when evaluating on NVIDIA and Dycheck iPhone to align with previous methods. 
% For static training, we train at $128\times128$ resolution for 90K iterations with learning rate $4e-4$ and batch size 256, then at $256\times256$ resolution for 90K iterations with learning rate $2e-4$ and batch size 256, and finally at $512 \times 512$ for 50K iterations with learning rate $1e-4$ and batch size 64. The training is done on 32 A100 GPU. For the dynamic stage, we train at $512\times512$ resolution for 10K iterations with learning rate $5e-5$ and batch size 128, and then increase the number of context views from 4 to 12 for 5K more iterations at learning rate $2.5e-5$ and batch size 64. The dynamic training is finished on 64 A100 GPUs. Note that since previous methods report results on NVIDIA and Dycheck iPhone at 270x480 resolution and 480x360 respectively, we change the resolution at the dynamic training stage to their specified resolutions instead of 512x512 for evaluation. 
We use the same training strategy for \enhancer{}. The numbers of iterations are 140K, 60K, and 30K for the progressive training in \textbf{Stage 1}, and are 20K for \textbf{Stage 2}, with the same learning rate schedule as above. 
% We borrow gradient clipping and skipping from~\cite{jin2024lvsm} to stabilize training.
% We use 4 supervision views in the training regardless of the number of input views. For static training, one of the supervision views is set to a randomly selected context view for easier convergence.
As introduced in \cref{sec:3d-pretrain}, we use a mixture of multiple datasets for training \cite{deitke2023objaverse,yu2023mvimgnet,zhou2018stereo,ling2024dl3dv,greff2022kubric,zheng2023point,karaev2023dynamicstereo,mehl2023spring} along with our 40K annotated dataset on PANDA-70M~\cite{chen2024panda}.
Note that we make sure that none of the testing scenes we show below is included in the training datasets. 
%\vspace{-2mm}

\parahead{Inference cost} Our model can be flexibly applied to different resolutions and numbers of context views.
% The inference time depends on the resolution and number of context views.
Measured on a single NVIDIA A100 GPU, \methodname{} takes \qty{20}{\milli\second} for 4-view $256^2$ reconstruction, \qty{150}{\milli\second} for the same resolution with 12 views, and \qty{1.55} {\second} for 12-view $512^2$ reconstruction. It requires less than 10 GB memory, which easily fits on a commercial-grade GPU (Result shown in Supplement). Please note that our model inference can be parallelized and the overall time overhead remains constant given sufficient memory. %and \qty{4.2}{\second} for 12-view $512\times896$ reconstruction. \enhancer{} takes \qty{0.44}{\second} seconds for 4-frame $512\times896$ reconstruction without KV caching. 
\begin{table*}%[t]
    \hspace{-2em}
\centering
\scriptsize
\begin{subtable}{0.4\textwidth}
\setlength{\tabcolsep}{2.5pt}
\begin{tabular}{lcccc}
\toprule
\textbf{Model} & \textbf{Rec. Time} & \textbf{PSNR}$\uparrow$ & \textbf{SSIM}$\uparrow$ & \textbf{LPIPS}$\downarrow$ \\ \midrule
TiNeuVox \cite{fang2022fast} & \qty{0.75}{\hour} & 14.03 & 0.502 & 0.538 \\
NSFF \cite{li2021neural} & \qty{24}{\hour} & 15.46 & 0.551 & 0.396 \\
T-NeRF \cite{gao2022monocular} & \qty{12}{\hour} & \cellcolor{first}16.96 & \cellcolor{first}0.577 & 0.379 \\
Nerfies \cite{park2021nerfies} & \qty{24}{\hour} & 16.45 & \cellcolor{second}0.570 & \cellcolor{third}0.339 \\
HyperNeRF \cite{park2021hypernerf} & \qty{72}{\hour} & \cellcolor{second}16.81 & 0.569 & \cellcolor{first}0.332 \\ \midrule
PGDVS \cite{zhao2024pseudo} & \qty{3}{\hour}$^{\dagger}$ & 15.88 & 0.548 & 0.340 \\
Depth Warp & -- & 7.81 & 0.201 & 0.678 \\ \midrule
\methodname{} (\textbf{Ours}) & \qty{0.98}{\second} & \cellcolor{third}16.52 & \cellcolor{second}0.570 & \cellcolor{second}0.338 \\
\bottomrule
\end{tabular}
\vspace{-1mm}
\caption{}
\label{tab:iphonebenchmark}
\end{subtable}
\hspace{3em}
\begin{subtable}{0.4\textwidth}
\setlength{\tabcolsep}{2.5pt}
\begin{tabular}{lcccc}
\toprule
\textbf{Model} & \textbf{Rec. Time} & \textbf{Render FPS} & \textbf{PSNR}$\uparrow$ & \textbf{LPIPS}$\downarrow$ \\ \midrule
HyperNeRF \cite{park2021hypernerf} & \qty{64}{\hour} & 0.40   & 17.60 & 0.367 \\
DynNeRF \cite{gao2021dynamic} & \qty{74}{\hour}  & 0.05 & \cellcolor{first}26.10 & \cellcolor{third}0.082 \\
RoDynRF \cite{liu2023robust} & \qty{28}{\hour} & 0.42  & \cellcolor{second}25.89 & \cellcolor{first}0.065 \\
4D-GS \cite{wu20244d} & \qty{1.2}{\hour} & 44 & 21.45 & 0.199 \\
Casual-FVS \cite{lee2025fast} & \qty{0.25}{\hour} & 48 & 24.57 & \cellcolor{second}0.081 \\ \midrule
PGDVS \cite{zhao2024pseudo} & \qty{3}{\hour}$^{\dagger}$ &     0.70    & 24.41 & 0.186 \\
Depth Warp & -- &  -- & 12.63 & 0.564 \\ \midrule
\methodname{} (\textbf{Ours}) & \qty{0.78}{\second} & 115 & \cellcolor{third}25.82 & 0.086 \\
\bottomrule
\end{tabular}
\vspace{-1mm}
\caption{}
\label{tab:nvidiabenchmark_new}
\end{subtable}
\vspace{-1mm}
\caption{\textbf{Quantitative comparisons on dynamic datasets.} (a) DyCheck iPhone dataset~\cite{gao2022monocular} comparison. (b) NVIDIA Dynamic Scene dataset~\cite{yoon2020novel} comparison. The results are rendered on $480\times 270$ resolution. `Rec. Time' is per-scene reconstruction time. $^\dagger$: Video-consistent depth estimation step included. We highlight the \colorbox{first}{best}, \colorbox{second}{second best}, and \colorbox{third}{third best} results.}
\label{tab:bothbenchmarks}
\vspace{-1.8em}
\end{table*}

\begin{figure}[ht!]
    \centering
    \includegraphics[width=\linewidth]{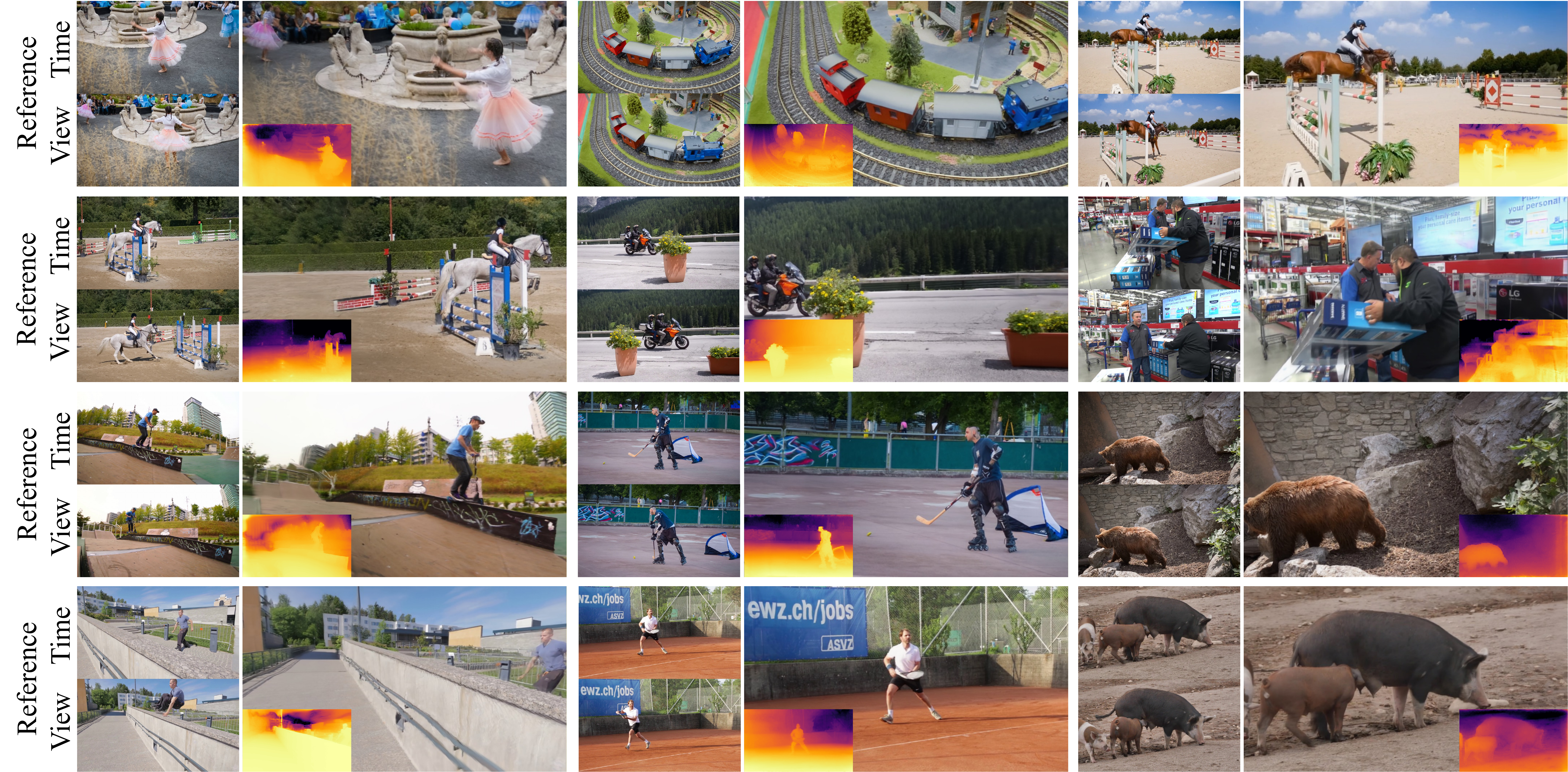}
    \caption{\textbf{Visualizations on DAVIS dataset~\cite{perazzi2016benchmark}.} We show our renderings on novel combinations of view poses and timestamps, with the correspondending references shown on the left. The lower-left/right corner shows the rendered depth map for each example.}
    \label{fig:davisresult}
    \vspace{-1em}
\end{figure}
\vspace{-1em}
\begin{figure}
    \centering
    %\vspace{-1em}
    \includegraphics[width=1.0\linewidth]{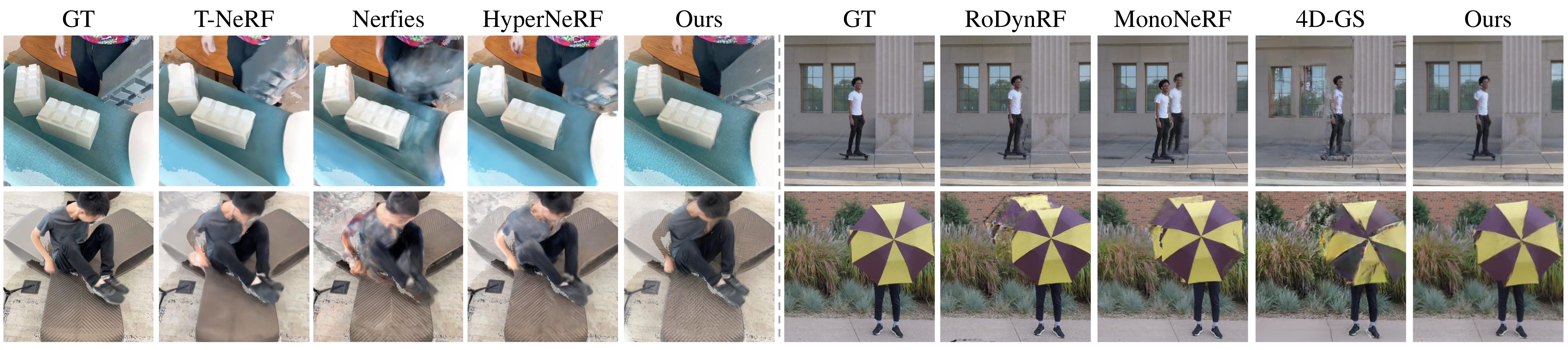}
    \caption{\textbf{Qualitative results} on DyCheck~\cite{gao2021dynamic} (left) and NVIDIA dynamic scene~\cite{yoon2020novel} (right) benchmarks.}
    \label{fig:qualitative_nsffiphone}
   \vspace{-2.0em}
\end{figure}
\subsection{Dynamic novel view synthesis}
\label{subsec:exp:dynamic}

%Comparison with MonST3R~\cite{zhang2024monst3r}.

\subsubsection{Quantitative analysis}

% To provide a fair comparison of our method with existing baselines, 
We provide quantitative evaluations on two of the largest dynamic view synthesis benchmarks.

\parahead{DyCheck benchmark~\cite{gao2022monocular}}%
The benchmark includes a dataset that contains 7 dynamic scenes recorded by 3 synchronized cameras. %: a hand-held iPhone and two stationary cameras positioned with a wide baseline. 
Following the protocol in \cite{gao2022monocular}, we take images from the iPhone camera as our context frames and use the frames from the 2 other stationary cameras for evaluation (totaling 3928 images of resolution $360\times480$). 
Our baselines include per-scene optimization-based methods, \ie, TiNeuVox~\cite{fang2022fast}, NSFF~\cite{li2021neural}, T-NeRF~\cite{gao2022monocular}, Nerfies~\cite{park2021nerfies} and HyperNeRF~\cite{park2021hypernerf}.
We additionally compare to a pseudo-feed-forward approach PGDVS~\cite{zhao2024pseudo}.
% To the best of our knowledge there is no published method that can be applied without per-scene optimization.
% We hence compare to direct depth warping and PGDVS~\cite{zhao2024pseudo}.
% The latter requires consistent depth estimation as input, which takes hours of GPU hours to estimate.

% We compare to scene-specific methods to assess whether per-scene
% optimization can be reduced. 
% These methods include TiNeuVox \cite{fang2022fast}, NSFF  \cite{li2021neural}, T-NeRF \cite{gao2022monocular}, Nerfies \cite{park2021nerfies} and  HyperNeRF ~\cite{park2021hypernerf}. For methods without per-scene appearance optimization, we compare against direct depth warping and PGDVS\cite{zhao2024pseudo}. Although PGDVS does not need per scene apperance optimization, it requires consistent depth estimation stage for each video which takes around three GPU hours. 

We report masked Peak Signal-to-Noise Ratio (PSNR), Structural Similarity Index Measure (SSIM)~\cite{wang2004image}, and LPIPS following the benchmark protocol~\cite{gao2022monocular} in \cref{tab:iphonebenchmark}, and show visualizations in \cref{fig:qualitative_nsffiphone}.
Note that since multi-frame inference can run in parallel, for our model we report single-frame reconstruction time regardless of video lengths.
It is encouraging to observe that even without per-scene optimization, \methodname{} achieves a very competitive performance compared to the baselines, ranking $2^\text{nd}$ in both SSIM and LPIPS scores. 
Our model surpasses PGDVS across all 3 metrics without the need of consistent depth estimate. %, achieving superior results without the need for an additional scene-fitting stage. 
This demonstrates our model's efficiency and strong generalization capability, being capable of providing sharper details and richer textures.
% Qualitative results in Figure~\ref{fig:qualitative_nsffiphone} also show that our model provides sharper and richer details in visible masked area. 

% \begin{figure}
%     \centering
%     \includegraphics[width=0.8\linewidth]{figs/ablation_static.pdf}
%     \caption{\textbf{Qualitative comparison of  models trained on different datasets} and evaluated on the out-of-distribution Tanks \& Temples benchmark~\cite{fan2024instantsplat}. }
%     \label{fig:ablation_static}
%     %\vspace{-1.0em}
% \end{figure}

\parahead{NVIDIA dynamic scene benchmark~\cite{yoon2020novel}}%
NVIDIA Dynamic Scene dataset contains 9 scenes captured by 12 forward-facing synchronized cameras. 
Following the protocol in DynNeRF~\cite{gao2021dynamic}, we build the input by selecting the frames at different timestamps in a `round-robin' manner. %\jhc{Is 'round-robin' a established term?}\HX{yes, it's been used in other baseline papers}
Then we evaluate the novel view synthesis quality at the first camera view but at different timestamps. We compare against HyperNeRF~\cite{park2021hypernerf}, DynNeRF~\cite{gao2021dynamic}, %NSFF~\cite{li2021neural}, 
RoDynRF~\cite{liu2023robust}, %MonoNeRF~\cite{tian2023mononerf}, 
4D-GS~\cite{wu20244d}, Casual-FVS~\cite{lee2025fast} as per-scene optimization baselines. 

Our results are shown in \cref{tab:nvidiabenchmark_new} and \cref{fig:qualitative_nsffiphone}.
% We show the visual comparisons
% in Fig.~\ref{fig:qualitative_nsffiphone} and an overall speed-quality comparison in Table.~\ref{tab:nvidiabenchmark}. 
Our model demonstrates performance that is competitive or exceeds that of previous optimization-based methods, ranking $3^\text{rd}$ among all baselines in terms of PSNR.
Compared to the explicit 3DGS-based representation~\cite{wu20244d,lee2025fast}, our approach outperforms their performance by 5\% on PSNR (25.82dB vs. 24.57dB).
% Our model ranks $3^\text{rd}$ among all methods in terms of PSNR, 
% showing sharper and richer details in dynamic content as demonstrated in Figure~\ref{fig:qualitative_nsffiphone}.
In terms of training and rendering speed, NeRF-based methods~\cite{gao2021dynamic,tian2023mononerf} require multiple GPUs and/or $>$1 day for optimization. 
Compared to \cite{wu20244d,lee2025fast}, our feed-forward bullet-time formulation is significantly faster, requiring no optimization time and rendering in real-time.
\vspace{-1.0em}
\begin{figure}[ht!]
    \centering
    \begin{subfigure}{0.56\linewidth}
        \centering
        \includegraphics[width=\linewidth]{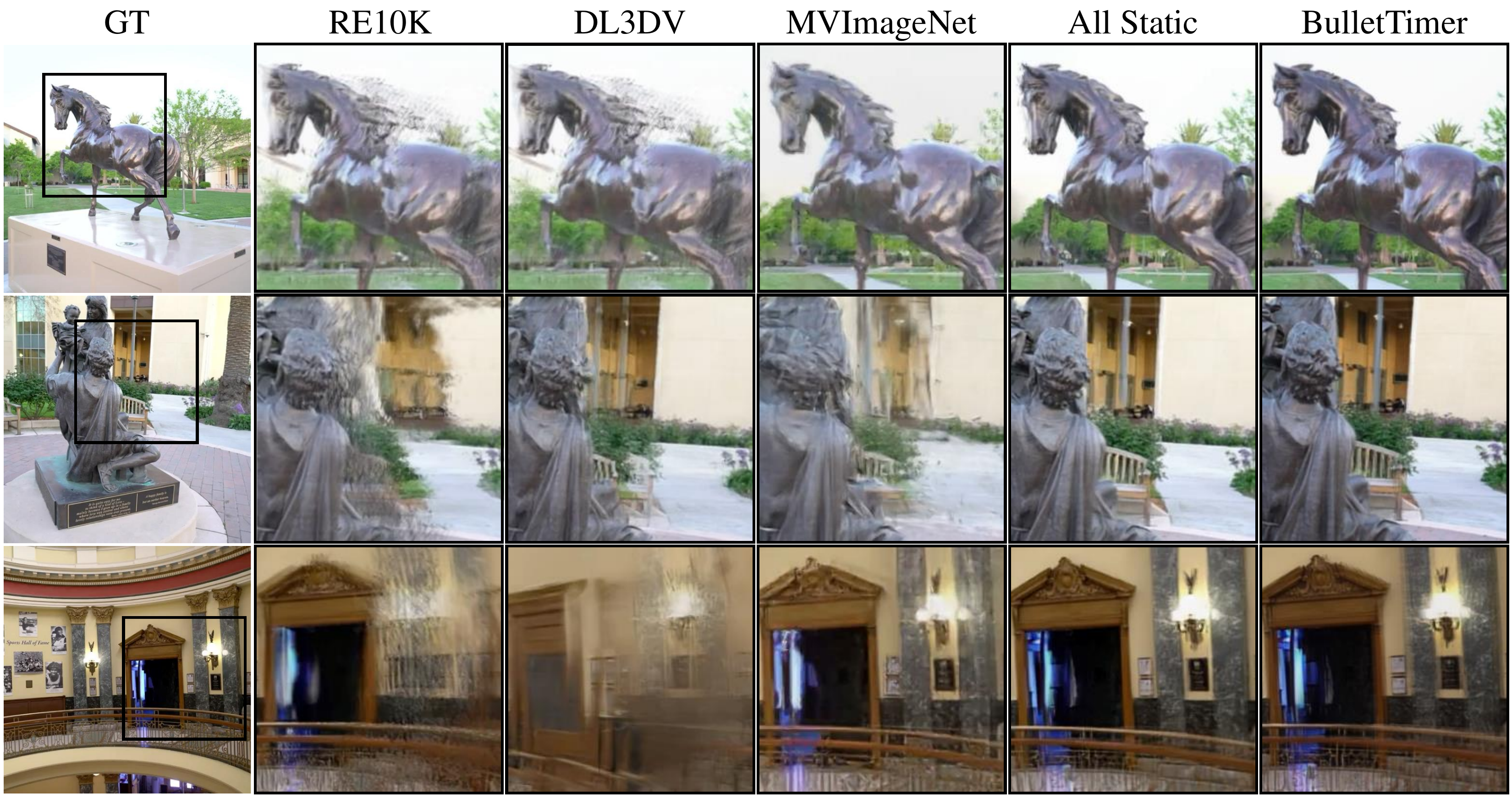}
        \vspace{-1.5em}
        \caption*{}
        \label{fig:ablation_static}
    \end{subfigure}
    \hfill
    \begin{subfigure}{0.43\linewidth}
        \centering
        \includegraphics[width=\linewidth]{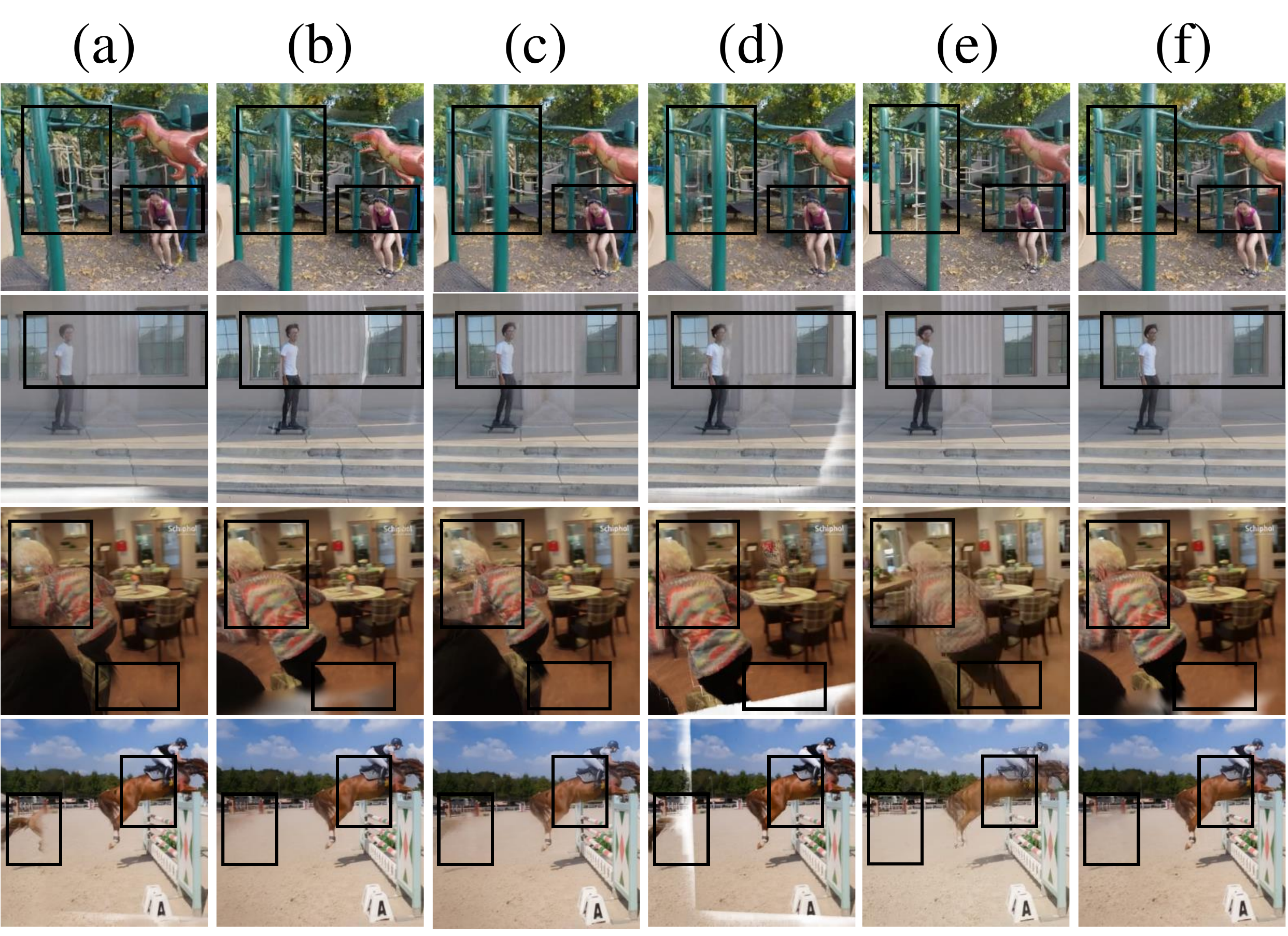}
        \vspace{-1.5em}
        \caption*{}
        \label{fig:abalation}
    \end{subfigure}
    \vspace{-1.5em}
    \caption{\textit{Left}: \textbf{Qualitative comparison of  models trained on different datasets} and evaluated on the out-of-distribution Tanks \& Temples benchmark~\cite{fan2024instantsplat}. \textit{Right}: (a) model w/o 3D Pretrain, (b) model w/ Re10K only 3D Pretrain, (c) model w/o static Co-train in \textbf{Stage 2}, (d) model w/o interpolation supervision, (e) Novel Time Enhancer model, (f) our full model. The upper two scenes are from NVIDIA dataset, and lower two scenes are from DAVIS dataset. }
    \label{fig:combined_ablation}
    \vspace{-2em}
\end{figure}

\subsubsection{Qualitative analysis}

% \subsubsection{Qualitative Results on the DAVIS  Datatset}\label{sec:davis}
To assess the performance of our method in real-world scenarios, we select multiple monocular videos from the DAVIS dataset~\cite{perazzi2016benchmark} for testing. 
% These videos encompass both indoor and outdoor settings, showcasing a wide range of objects, including animals, humans, toys, various structures, and intricate geometries like flamingo legs and rails in the House-Jump scene. 
% The scenarios present diverse types of motion such as hockey play, dancing, walking, loading, and jumping, along with challenging geometries that introduce frequent occlusions and complex dynamics. 
% These features pose significant challenges for 4D reconstruction models. 
% Camera poses for the video frames were estimated using xxx, which takes 0.2s per frame\HX{@JIAHUI, could you please add more details}. 
Camera poses for the videos were estimated using the same annotation technique as detailed in \cref{sec:3d-pretrain}.
% As discussed in Section \ref{inference}, to reconstruct the scene at a specific target time, we first identify the frame nearest to the target time, followed by uniformly selecting an additional 15 frames, which together serve as the RGB input to our model.
\cref{fig:davisresult} shows a visualization of the results. 
% For each scene, we illustrate renderings from a target view at a specified target time, along with the corresponding video frames for reference. 
Our model demonstrates strong generalization capabilities in real-world captures, producing high-quality, sharp renderings across a variety of objects with complex motions while maintaining robust temporal and multiview consistency. 

\begin{wrapfigure}{r}{0.5\textwidth}
\vspace{-1.0em}
\begin{minipage}{0.5\textwidth}
\centering
\scriptsize
\begin{subfigure}{0.45\textwidth}
\setlength{\tabcolsep}{1.5pt}
\begin{tabular}{lc}
\toprule
\textbf{Model} & \textbf{LPIPS}$\downarrow$ \\ \midrule
GPNR~\cite{suhail2022generalizable} & 0.250 \\
PixelSplat~\cite{charatan2024pixelsplat} & 0.142 \\
MVSplat~\cite{chen2025mvsplat} & 0.128 \\
GS-LRM~\cite{zhang2025gs} & \cellcolor{third}{0.114} \\ \midrule
\textbf{Ours}-Static & \cellcolor{first}{0.070} \\ 
\midrule
\textbf{Ours}-Full & \cellcolor{second}{0.089} \\
\bottomrule
\end{tabular}
\caption{}
\label{tab:staticre10k}
\end{subfigure}
\hspace{-1.5em}
\begin{subfigure}{0.52\textwidth}
\setlength{\tabcolsep}{1.pt}
\begin{tabular}{llc}
\toprule
\textbf{Model} & \textbf{Datasets} & \textbf{LPIPS}$\downarrow$ \\ \midrule
GS-LRM$^{*}$~\cite{zhang2025gs} & RE10K & 0.310 \\ \midrule
\multirow{4}{*}{\textbf{Ours}-Static} & Objaverse & 0.668 \\
 & MVImageNet & 0.343 \\
 & DL3DV & \cellcolor{third}{0.278} \\
 & All Static & \cellcolor{first}{0.093} \\ \midrule
\textbf{Ours}-Full & +Dynamic & \cellcolor{first}{0.093} \\
\bottomrule
\end{tabular}
\caption{}
\label{tab:staticinstantsplat}
\end{subfigure}
\caption{\textbf{Quantitative comparisons on static datasets.} (a) results on the RE10K benchmark~\cite{zhou2018stereo}; (b) results on the Tanks and Temples benchmark~\cite{fan2024instantsplat}. We highlight the \colorbox{first}{best}, \colorbox{second}{second best}, and \colorbox{third}{third best} models. $^{*}$: Our reproduced results.}
\label{tab:staticbaseline}
\end{minipage}
\vspace{-1.5em}
\end{wrapfigure}
% \JR{Since we don't have results  on dynamic-to-static evaluation results yet, I moved the full static evaluation to supp, and leave a concise version in Ablation (Effect of Cross-dataset Static Training).}

\subsection{Compatibility with static scenes}
\label{subsec:exp:static}

Although our model is primarily designed to handle dynamic scenes, the formulation and the training strategy enable it to be still backward compatible with static scenes. In this section, we show that the \emph{same} model achieves competitive results on static scenes.
% Interestingly, our bullet-time reconstruction model, originally intended for dynamic scene reconstruction, demonstrates strong generalization ability with impressive performance on out-of-distribution static dataset. While this is not our main contribution, we view it as a bonus of our design.

\parahead{RealEstate10K (RE10K) benchmark~\cite{zhou2018stereo}} 
% In line with prior work on static reconstruction from extremely sparse inputs, 
We evaluate our model on the RE10K dataset and compare with several state-of-the-art models~\cite{suhail2022generalizable,charatan2024pixelsplat,chen2025mvsplat,zhang2025gs}. To ensure comparability with baseline models, we train and test our model using $256\times256$ resolution. \cref{tab:staticre10k} presents a quantitative comparison on LPIPS, where our static model outperforms all the baselines.  %While our original pre-trained static model takes 4 context views as input, we fine-tune it with only 2 context views for 10K iterations for a fair comparison. \cref{tab:staticre10k} presents a quantitative comparison on LPIPS, where both our original 4-view static model and the fine-tuned 2-view static model outperform baselines. 
% Notably, although primarily as a dynamic scene reconstruction model, \methodname{} can produce highly competitive results with better LPIPS scores than all previous baselines. 
Please refer to the Supplement for more comparisons on other metrics and visualizations.
% \cref{tab:staticre10k} presents a quantitative comparison on LPIPS, demonstrating that our static model consistently outperforms the baselines. Our original static model, which accommodates additional context views, achieves even greater performance improvements. Notably, our dynamic model can produce decent results with better perceptual scores than all previous baselines. Please refer to Supplementary for a comparison on PSNR and LPIPS scores as well as a qualitative comparison.

\parahead{Tanks \& Temples benchmark~\cite{fan2024instantsplat}} 
We further evaluate our model on an unseen test dataset, the Tanks \& Temples~\cite{knapitsch2017tanks} subset from the InstantSplat~\cite{fan2024instantsplat} benchmark, which consists of 10 scenes. 
We use the state-of-the-art novel view synthesis model~\cite{zhang2025gs} as our baseline, reproducing their model since the original code and weights are not publicly available. 
Additionally, we include our pretrained static model from \textbf{Stage 1} as an additional baseline. 

To analyze the impact of our mixed-dataset pretraining strategy, we also train single-dataset models using the same training schedule as further baselines. All models utilize 4 context views. Quantitative results (\cref{tab:staticinstantsplat}) demonstrate that our pretrained static model with mixed-dataset training substantially outperforms the single-dataset models, highlighting the crucial role of multi-dataset training for generalization to unseen domains. Even when incorporating the dynamic scene datasets, \methodname{} achieves comparable result to our best static models. \cref{fig:ablation_static} provides a qualitative comparison, showing that \methodname{} consistently generates sharper outputs that closely align with the ground truth.
\vspace{-0.5em}

\subsection{Ablation study}
\label{sec:ablation}
\begin{figure}[t]
    \centering
    \begin{subfigure}{0.29\linewidth}
        \centering
        \includegraphics[width=\linewidth]{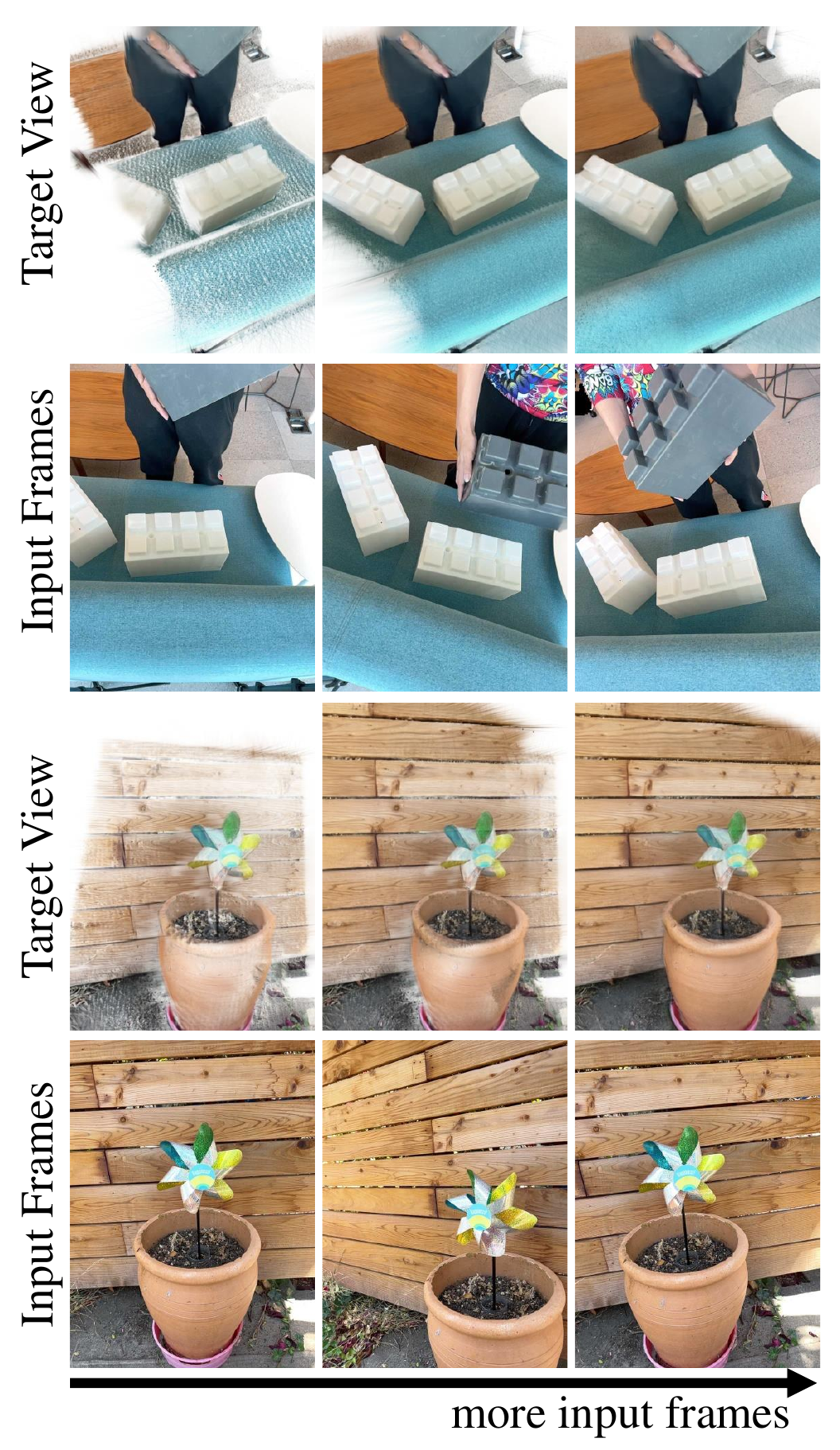}
        \vspace{-1.5em}
        \caption{}
        \label{fig:bullettime}
    \end{subfigure}
    \hfill
    \begin{subfigure}{0.7\linewidth}
        \centering
        \includegraphics[width=\linewidth]{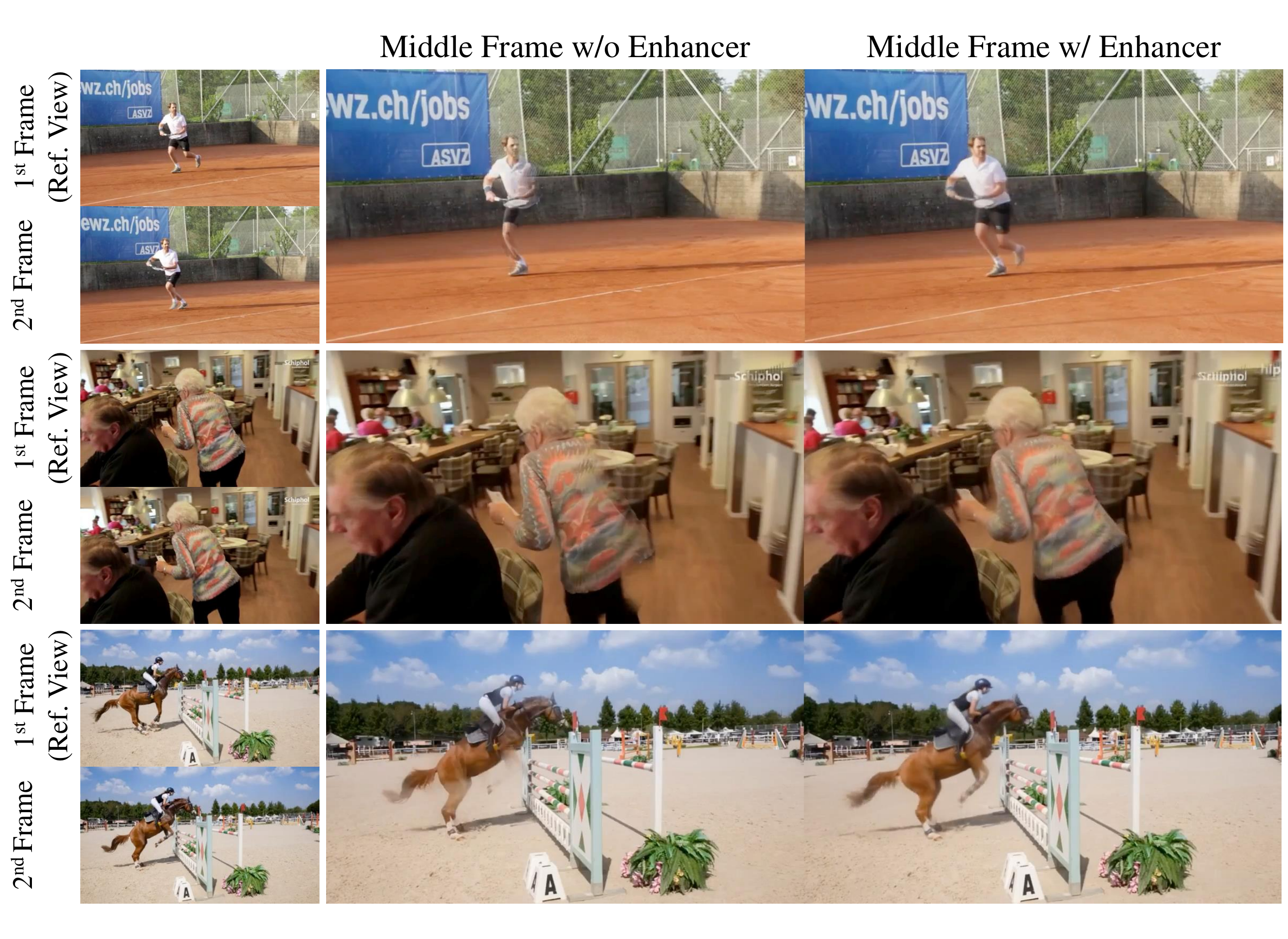}
        \vspace{-1.5em}
        \caption{}
        \label{fig:ablation_nte}
    \end{subfigure}
    \vspace{-1.5em}
    \caption{ (a) \textbf{Illustration of bullet-time reconstruction from multiple context frames.} Increased
number of frame predictions leads to progressively more
complete scene reconstruction on target views. (b) \textbf{Ablation on the \enhancer{} module}. The middle frame is in between the 1$^\text{st}$ frame and the 2$^\text{nd}$ frame. Results are rendered from the view of the 1$^\text{st}$ frame.}
    \label{fig:combined_bullet_ablation}
    \vspace{-1.5em}
\end{figure}

We study the effect of different design choices. \textbf{1) Context frames.}
We visualize the reconstruction results as we progressively add 3DGS predictions from more context frames across multiple different timestamps in \cref{fig:bullettime}, where increasing the number of context frame leads to progressively more complete scene reconstruction. This demonstrates the flexibility of our bullet-time reconstruction formulation: during the inference stage, we can arbitrarily select spatially-distant frames that contribute to a more complete view coverage of the scene.
\textbf{2) Curriculum training.}
We show in \cref{fig:abalation} the effect of our curriculum training strategy.
Without \textbf{Stage 1} of pre-training on static scenes, the model struggles to produce results of correct geometry and sharp details.
Pretraining on multiple diverse datasets is also crucial, which we demonstrate by \emph{just} training on RE10K dataset, and non-negligible distortions are observed in the results.
Similarly, even in \textbf{Stage 2} of our curriculum, we still need to co-train on static scenes which provide more multi-view supervisions, thus maintaining the rich details and reasonable geometries. Quantitative ablation results are shown in the Supplement.
\textbf{3) Interpolation supervision.}
Shown in \cref{fig:abalation} (with more results in the Supplement), interpolation supervision (introduced in \cref{sec:bullet-time model}) plays a significant role, without which the model tends to produce \emph{white-edge} artifacts.
This occurs because without interpolation loss, the model often generates 3DGS that are positioned too close to the camera with low depth values to \emph{cheat} the loss.
In contrast, adding the interpolation supervision requires the model to account for scene dynamics and encourages consistency across multiple views. 
\textbf{4) \enhancer.}
As demonstrated in \cref{fig:ablation_nte}, our \enhancer{} module enhances the bullet-time reconstruction model's ability to handle scenes with fast or complex motions, largely reducing the ghosting artifacts. 
Additional video results are provided in the supplementary material. Although 3D-free design enables NTE to handle complex dynamics and produce smooth transitions between adjacent frames, the model struggles to produce novel views that are far from the input camera trajectory (As illustrated in \cref{fig:abalation}). \vspace{-1.0em}

\section{Conclusion}
\label{sec:conclusion}
\vspace{-1.0em}
In this paper we present \methodname{}, the first feed-forward dynamic 3D scene reconstruction model for novel view synthesis.
We present a bullet-time formulation that allows us to train the model in a more flexible and scalable way.
We demonstrate through extensive experiments that our model is able to provide high-quality results at arbitrary novel views and timestamps, outperforming the baselines in terms of both quality and efficiency.

\parahead{Limitations} Our method is mainly targeted for novel view synthesis, and the recovered geometry (hence the depth map) is usually not as accurate. %as the most recent depth prediction models (\eg \cite{yang2024depth}). 
Correspondences between frames are implicitly modeled by the neural network, and our pixel-aligned Gaussian representation cannot represent temporal deformations.
Although practically we observe temporally coherent results, additional post-processing steps have to be introduced to recover the explicit motion of the geometry.
% The pixel-aligned representation restricts BTimer from learning deformations. Post-processing might be required to ensure temporal smoothness since frames are reconstructed individually.
% Our model also has limited support for view \emph{extrapolation}.
% Incorporating a generative prior in the loop is a promising direction to pursue in the future.

\textbf{Broader Impact.} \methodname{} can transform posed casual videos into realistic dynamic 3D assets. However, it should be used with caution, particularly concerning privacy, copyrights, and the potential for malicious impersonation.
% TODO: add discussion about limitation of dynamics. Add offsets in the supplemental.

%\section*{References}

%References follow the acknowledgments in the camera-ready paper. Use unnumbered first-level heading for the references. Any choice of citation style is acceptable as long as you are consistent. It is permissible to reduce the font size to \verb+small+ (9 point) when listing the references. Note that the Reference section does not count towards the page limit. \medskip
%{
%    \small
    % \bibliographystyle{ieeenat_fullname}
   % \bibliography{main}
%}
\bibliographystyle{unsrt}
\bibliography{main}

%%%%%%%%%%%%%%%%%%%%%%%%%%%%%%%%%%%%%%%%%%%%%%%%%%%%%%%%%%%%

%\appendix

%\section{Technical Appendices and Supplementary Material}
%Technical appendices with additional results, figures, graphs and proofs may be submitted with the paper submission before the full submission deadline (see above), or as a separate PDF in the ZIP file below before the supplementary material deadline. There is no page limit for the technical appendices.

%%%%%%%%%%%%%%%%%%%%%%%%%%%%%%%%%%%%%%%%%%%%%%%%%%%%%%%%%%%%
%\newpage

%\begin{center}
%    \vspace*{1.5em}
%    \textbf{\LARGE Supplementary Material}
%    \vspace*{0.5em}
%\end{center}
%\input{neurips_sec/X_suppl}
\end{document}